# Detection, Recognition and Tracking of Moving Objects from Real-time Video via SP Theory of Intelligence and Species Inspired PSO

Kumar S. Ray, Sayandip Dutta, Anit Chakraborty

***Abstract*— In this paper, we address the basic problem of recognizing moving objects in video images using SP Theory of Intelligence. The concept of SP Theory of Intelligence which is a framework of artificial intelligence, was first introduced by Gerard J Wolff, where S stands for Simplicity and P stands for Power. Using the concept of multiple alignment, we detect and recognize object of our interest in video frames with multilevel hierarchical parts and subparts, based on polythetic categories. We track the recognized objects using the species based Particle Swarm Optimization (PSO). First, we extract the multiple alignment of our object of interest from training images. In order to recognize accurately and handle occlusion, we use the polythetic concepts on raw data line to omit the redundant noise via searching for best alignment representing the features from the extracted alignments. We recognize the domain of interest from the video scenes in form of wide variety of multiple alignments to handle scene variability. Unsupervised learning is done in the SP model following the 'DONSVIC' principle and 'natural' structures are discovered via information compression and pattern analysis. After successful recognition of objects, we use species based PSO algorithm as the alignments of our object of interest is analogues to observation likelihood and fitness ability of species. Subsequently, we analyze the competition and repulsion among species with annealed Gaussian based PSO. We've tested our algorithms on David, Walking2, FaceOcc1, Jogging and Dudek, obtaining very satisfactory and competitive results.**

## I. Introduction

Effective recognition of objects and tracking of the recognized objects in a video scene in real-time video stream is a very challenging task for any visual surveillance systems. In this paper, we perform several tasks, such as: extraction of multiple alignments, parsing of raw data line to obtain best noise free alignment for more accurate recognition and obtaining optimum solution via family resemblance or polythetic concept, analyzing a scene with high-level feature alignments recorded with more logical detailing about its existence in the raw data. Finally, we track the recognized object of interest using species inspired Particle Swarm Optimization (PSO).

In the past, there have been many attempts to achieve human like perception or to handle Computer Vision related problems strictly in a logical manner, i.e. using atomic symbols instead of using actual numerical data. Default logic based reasoning [13], [14] and bilattice based non-monotonic reasoning [25] have been applied in the field of visual surveillance. But it sometimes generates unexpected extensions. Conclusions drawn from default logic vindicates common sense, which in turn jeopardizes its soundness. Other logic based reasoning systems include Neutrosophic logic [4], which aims to improve on the basic prospects of fuzzy logic. However, it can often get paradoxical and sometimes unintuitive results. In these cases, SP systems provide a more simple and robust framework to interpret logic, which is ideal for problem solving in the domain of computer vision. In order to track the recognized objects of interest we have used species inspired Particle Swarm Optimization technique [19]. The multiple alignments and the class and subclass hierarchies that are derived from SP systems are analogous to the species based framework we have used in our tracking approach. Thus, unlike conventional state-of-the-art PSO algorithms [1], [2], [3], [8], [9], [12], [17], [22] species based PSO technique is more well equipped to process the multiple alignments generated from the SP framework.

In our experiment, we use natural language texts of multiple alignments for SP systems to extract necessary information from raw data line in form of Old pattern, and comprehend the knowledge base in test domain to derive and encode New information. The system, in turns, learns from its knowledge base exploring wide variety of alignments to create and compress information to form New relevant patterns, without any supervision via DONSVIC principle. Through its thorough experiences from the knowledge base, the system is capable of exploring objects, class of objects from images and form necessary patterns to update the Old information with New. Using DONSVIC principle and polythetic concept, SP systems are well equipped to handle partial occlusion by searching for relevant pattern from its Old knowledge base.

Kumar S. Ray is with Indian Statistical Institute, 203 B.T Road, Kolkata 108, India. (e-mail: ksray@isical.ac.in)
Sayandip Dutta is with Indian Statistical Institute, 203 B.T Road, Kolkata 108, India. (e-mail: sayandip199309@gmail.com)
Anit Chakraborty is with Indian Statistical Institute, 203 B. T Road, Kolkata – 108, India. (e-mail: ianitchakraborty@gmail.com)

Furthermore, the optimum compressed pattern representing the knowledge base of our object of interest is tracked via species inspired PSO in successive frames.

The contributions of this paper are:

- Development of the concept of simplicity with descriptive or explanatory power in the field of Computer Vision.
- Development of brain-like visual inference system and its application in tracking objects from real-time video.
- Versatility and flexibility of artificial systems by means of unsupervised learning, planning, pattern analysis to attain seamless integration of artificial vision with other sensory modalities.
- Occlusion and redundant noise handling from the raw data line for accurate detection and recognition for tracking.
- Accurate detection of the dominant object between two overlapping species for more persistent tracking of the object of interest.

The organization of the paper constitutes: Review and overview of SP theory in Section II. Section III explains our proposed method for detection, recognition and tracking via SP systems and spices based PSO. Experimental observations are presented in Section IV. Section V concludes the paper and discuss future possibilities for further improvements in the field of Computer Vision and Machine Intelligence.

## II. Outline of SP Theory

Simplification and integration of concepts in cognition and computation can be achieved with the help of SP Theory of Intelligence with information compression as the overlying theme. This section briefly explains the basic concept of SP theory and some of its relevant application in the area of computer vision and natural vision, specifically, the basic problem of accurate detection, recognition of moving objects in video images and occlusion handling with significant advantages.

In broad terms, the SP theory has three principle elements:

- Knowledge is represented with patterns: One or two dimensional arrays of atomic symbols.
- Information is compressed by pattern matching and unification (merging) concept via Multiple Alignment, as demonstrated in Fig. 1.
- The learning of the system is achieved by compression of New patterns to create Old patterns and unification of alignments.

The system learns by compressing New patterns to create Old patterns like those shown in columns 1 to 11 in Fig. 1. Because of the reliant connection in between information compression and concepts of probability and prediction [32], the SP system is intrinsically probabilistic. Frequency of occurrences of the best possible multiple alignment is intimately associated with its subsequent SP patterns. Probabilities can be formulated for each associate inferences of subsequent multiple alignments [26]. It is intended that the SP computer model will be the basis for the development of a high-parallel SP machine, an expression of the SP theory, a vehicle for research, and a means for the theory to be applied [15].

Although the main emphasis in the SP program has been on the development of abstract concepts in natural language processing, but its application in computer vision and video image processing is being currently explored by different researchers.

### A. *The Multiple Alignment Concept*

Figure 2 demonstrates an example of multiple alignment in the SP system, where a sentence: `t w o k i t t e n s p l a y' is represented as a New pattern in row 0, whereas Old patterns are represented in rows 1 to 8 as a word with grammatical markers or a grammatical rule. This multiple alignment, parses the sentence in terms of grammatical structures or series of consecutive grammatical markers. This Multiple Alignment is the best of more than a few built by the SP62 model when it is provided with the New pattern and a series of consecutive grammatical markers of Old patterns that contains those shown in the figure and many others. In our example, `best' corresponds to the most economically encoded New pattern in terms of the Old patterns [26].

The `Np' and `Vp' marks the inter grammatical dependency of the plural subject (`k i t t e n s') and the plural main verb (`p l a y') of the sentence, in terms of a Multiple Alignment Point of Interest (MAPI). There may exist a discontinuous dependency between one element and the another. The term `discontinuous' represents the presence of large amounts of arbitrary intervening structures. Discontinuous dependency marking is, conceivably, more well-designed and easier than other state-of-the-art grammatical systems.

### B. *The encoding of light intensities*

Expressing the that light intensities in images as numbers is trivial while designing a Machine Learning and artificial systems for Computer Vision. But, the SP systems only recognizes atomic symbols of consecutive grammatical markers where every multiple alignment are matched with another in an all-or-nothing manner. In principle, it may interpret numerical values correctly if the machine is supplied with certain patterns that hold information that are similar to Peano’s axioms [31]. Although, this has not yet been explored in the research areas of Computer Vision and Machine Intelligence, nevertheless, numerical values are not the best way to assess the principle of SP Theory of Intelligence.

Initially, for simplicity, we assume that all the images are in Binary, i.e. pixel values are either ‘0’ or ‘1’. In such cases, the illumination variation of unit pixels at any given area of the image will be encoded as the distribution of pixel-intensities of Black and White in that area. This representation, somewhat, avoids the explicit numerical values of the corresponding pixels similar to dark and light monochrome photographs of old newspapers [27]. SP Systems welcomes the idea of atomic representation of unit pixel values as ‘0’ or ‘1’, without any numerical meanings.

```
0   1    2     3    4    5     6    7     8    9     10   11

         S
         Num -------------------------------------------- Num
                                                          PL
         ; ---------------------------------------------- ;
         NP ------------ NP
                         0
                    N -- N
                    Np ---------------------------------- Np
                    7
w ----------------- w
i ----------------- i
n ----------------- n
d ----------------- d
s ----------------- s
                    #N - #N
                         Q -------- Q
                                    1
                              P -- P
                              15
f ---------------------------- f
r ---------------------------- r
o ---------------------------- o
m ---------------------------- m
                              #P - #P
                                    NP ------- NP
                                               0a
                                               D --- D
                                                     17
t ---------------------------------------------------- t
h ---------------------------------------------------- h
e ---------------------------------------------------- e
                                               #D -- #D
                                          N -- N
                                          Ns
                                          4
w ---------------------------------------- w
e ---------------------------------------- e
s ---------------------------------------- s
t ---------------------------------------- t
                                          #N - #N
                                    #NP ------ #NP
                         #Q ------- #Q
         #NP ----------- #NP
         V --- V
               Vp --------------------------------------- Vp
               11
a ------------ a
r ------------ r
e ------------ e
         #V -- #V
    A -- A
    19
s - s
t - t
r - r
o - o
n - n
g - g
    #A - #A
         #S

0   1    2     3    4    5     6    7     8    9     10   11
```

Fig. 1. Multiple alignment and information compression by pattern matching. [26]

```
0                           t w o                 k i t t e n   s                  p l a y       0
                            | | |                 | | | | | |   |                  | | | |
1                           | | |          < Nr 5 k i t t e n > |                  | | | |       1
                            | | |          | |                | |                  | | | |
2                           | | |   < N Np < Nr               > s >                | | | |       2
                            | | |   | | |                         |                | | | |
3                  < D Dp 4 t w o > | | |                         |                | | | |       3
                   | |            | | | |                         |                | | | |
4             < NP < D            > < N |                         > >              | | | |       4
              | |                       |                           |              | | | |
5             | |                       |                           |       < Vr 1 p l a y >     5
              | |                       |                           |       | |            |
6             | |                       |                           | < V Vp < Vr          > >   6
              | |                       |                           | | | |                  |
7 < S Num   ; < NP                      |                           > < V |                  > > 7
       |    |                           |                                 |
8     Num PL ;                          Np                                Vp                     8
```

Figure 2. Demonstration of Grammatical Markers in SP system.
New pattern is represented in row 0 and Old patterns are represented in rows 1 to 8. [26]

### C. *Edge detection with the SP system*

Recursion can simulate the outcome of run-length coding in the SP framework, as demonstrated in Fig. 3. Here, each instance of self-referential Old pattern, in the form of 'X 1 a b c X #X #X', in rows 1 to 4, is matched from row 0, which contains each appearance of 'a b c' in New pattern. The enclosing 'X #X' in the body of the pattern can be unified and matched with 'X ... #X' at the beginning and end. Due to this property, the structure of Old pattern is called self-referential.

Derivation of relatively shorter multiple alignment sequence 'X 1 1 1 1 #X' is encoded from the New pattern. The recording of the fact that pattern 'a b c' contains 4 instances, attains the lossless compression of the initial original sequence by unary arithmetic. Recording a sequence of instances of 'a b c', irrespective of the length of the initial sequence, may reduce the encoding to 'X #X' with lossy compression. As briefly mentioned earlier in this paper, two side-by-side consecutive encodings, would be a uniform economical boundary between one subsequent region to another.

At an abstract level, there may exist two set of similar productions as an outcome: Redundancy of uniform regions is extracted from the raw data, without any careful consideration of boundaries between subsequent regions as an economical depiction of the raw data, as mentioned by 'primal sketch' [36]. Moreover, SP concepts are generalized to two dimensions, as a tool to attain significant breakthroughs in the field of Computer Vision and Machine Learning Systems.

### D. *Orientations, Lengths, and Corners*

In principle, the orientations of edges or their lengths may be mathematically encoded, very economically, with the help of vector graphics representation. Having said that, the aforementioned method may not be as useful for systems like: Molecular Biological Systems, Gene Technology etc. Also, in real life, it is very difficult to attain human like capabilities in an artificial perception system following vector graphical method.

As briefly mentioned earlier, in natural vision, quite simply the edges may be directly encoded either by matching neuron type or by multiple alignment based artificial systems. [26]

The orientation and length of a straight line may be obtained through sequence of codes containing significant amount of redundancy, as briefly mentioned earlier in Section II(C). Orientation of the sequence is repeated in succeeding parts of the raw line data. So, it is fair to assume, in natural vision and systems, redundancy is reduced with some run-length coding inside the body of the line. When repetition stops, the information is preserved at the beginning and ending points of the raw line data. [26]

This method is susceptible to straight lines as well as uniform curvature. Such structures, either partially or its repeated instances are encoded to express the curvature of the entire line.

Relevant information regarding the presence of 'end stopped' hypercomplex cells that are selectively responsive to a corner or a bar of a definite length, can be extracted with regard to straight lines [20]. It is safe to assume that, in mammalian vision, the length of an edge and the orientation, line or slit, is majorly encoded via edge detection using neurons to record the end to end point associated corners. The input line for a 'higher' level of encoding is provided via orientation-sensitive neurons.

In terms of artificial systems, in principle, this kind of approach is adapted within the means of multiple alignment framework as mentioned in Section II (A).

```
0     a b c     a b c     a b c     a b c                 0
      | | |     | | |     | | |     | | |
1     | | | X 1 a b c X   | | |     | | |         #X #X   1
      | | | |         |   | | |     | | |         |  |
2 X 1 a b c X         |   | | |     | | |         |  #X #X 2
                      |   | | |     | | |         |
3                     X 1 a b c X   | | |      #X #X      3
                                |   | | |      |
4                               X 1 a b c X #X #X         4
```

Figure 3. Edge detection with SP systems.
Each instance of self-referential Old pattern, in the form of ‘X 1 a b c X #X #X’, in rows 1 to 4, is matched from row 0, which contains each appearance of ‘a b c’ in New pattern. [26]

### *E. Noisy data and low-level features*

Visual data collected from raw images is hardly as clean as demonstrated above in Figure 4. Monochromic images likely to be carrying various kinds video frame impurities, such as: not purely black or purely white, shade of grey, and there are likely to be blots and smudges of various kinds. SP Systems are designed to search for optimal solutions and is not destructed by errors of commission, substitution and omission. There is more on this topic in Sections III (A, B).

## III. Proposed method

To successfully track objects across multiple frames, accurate detection of the objects of interest are of tremendous importance.

The primary objective of SP model is to figure out good full or partial match between different patterns with high efficiency, much like the standard models that are based on ‘dynamic programming’ for the sequence matching or alignment. However, the difference between the SP theory and the latter is that the former (SP model) delivers all the matching alternatives within the patterns; whereas the standard models (based on Dynamic Programming) are programmed only for the best solution. At every stage, multiple alignments are built by pairwise matching and unification of the patterns. The objective of this process is to encode New information in terms of Old information economically, so as to separate the subpar multiple alignments that are generated.

Despite the straightforwardness of the SP patterns, they are very much versatile in representing various kinds of knowledge, due to their processing within the multiple alignment frameworks.

This allows SP systems to process information such as, natural language grammar objects, part-whole hierarchies, class hierarchies, ontologies, if-then rules, relation tuples, decision trees, associations of medical symptoms with medical signs, causal relationships, and mathematical and logical concepts.

The SP system shows definite potential in the areas of natural pattern recognition, language processing, reasoning and inference frameworks, the efficient storage, compression and retrieval of information and unsupervised learning. As the leniency in multiple alignment process allows to filter out noisy and erroneous data, SP theory is quite robust in the face of errors. In this paper, we apply these traits to overcome the scenarios when an object is partially occluded from the camera viewpoint.

After successful detection and recognition of the object of interests we track them using species inspired Particle Swarm Optimization. This approach is very well equipped for processing multiple alignment and hierarchical data generated from the SP theory, which aids in a more persistent tracking of multiple moving objects of interests.

### *A. Object Detection and Recognition from Training Images*

Object recognition, in some respect, is similar to parsing in natural language processing [16], [18], [28]. SP system is quite well equipped with parsing natural language, as outlined in Section II (A), thus it can be considered as a useful tool for the development of Computer Vision and Pattern Recognition areas. Logically, SP machine needs to be generalized for working with patterns with two dimensions. In our experiment, though, we would consider the system to be well equipped to detect and identify low level perceptual features, which are initially atomic in nature to balance the harmony with the SP theory.

To put in perspective, Fig. 4 demonstrates schematically how a person’s face with all its atomic feature symbols [e.g. Ears, Nose, Eyes etc.] are parsed within the multiple alignment grammar. The New pattern, represented in row 0, contains the incoming information from the raw line data.

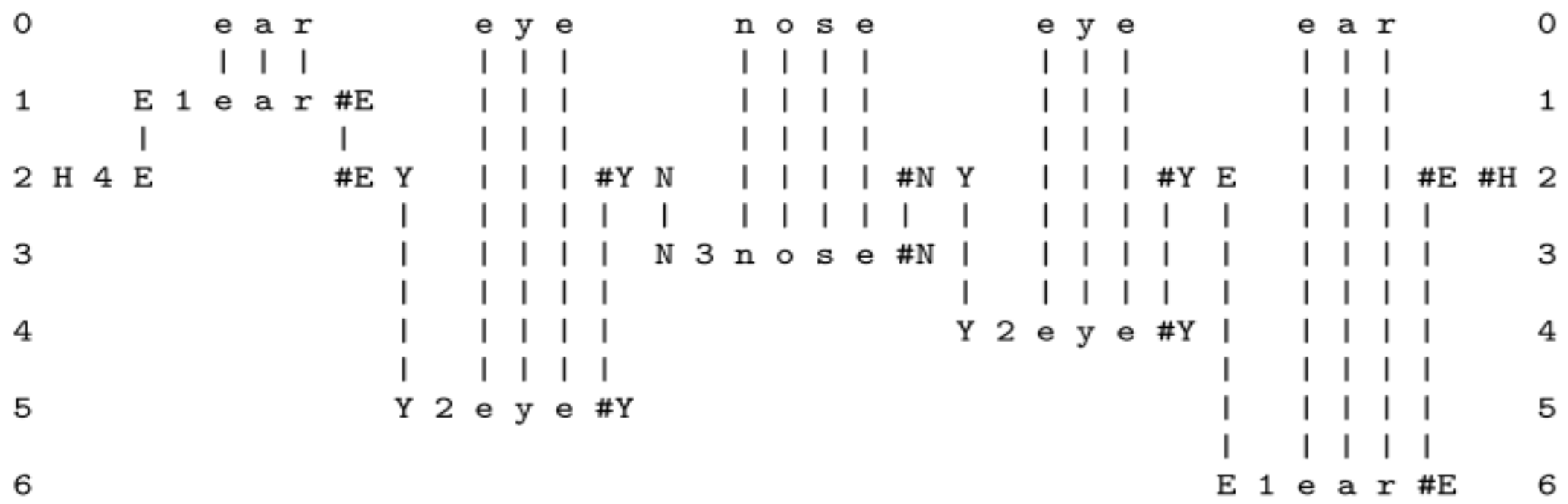

Figure 4. Schematic description of a person's face.
Its atomic symbols are parsed within the multiple alignment grammar. [26]

The stored knowledge of the structure of Ears, Nose, Eyes etc. is depicted in the Old patterns are aligned with every atomic feature of the object of interest. The updated multiple alignment is then matched with a pattern in row 2 as a relatively superior unit-feature of the object (i.e. Someone's head). Even though this method is schematic in nature, this approach has strong potential in our experiment, as explained in subsequent sections.

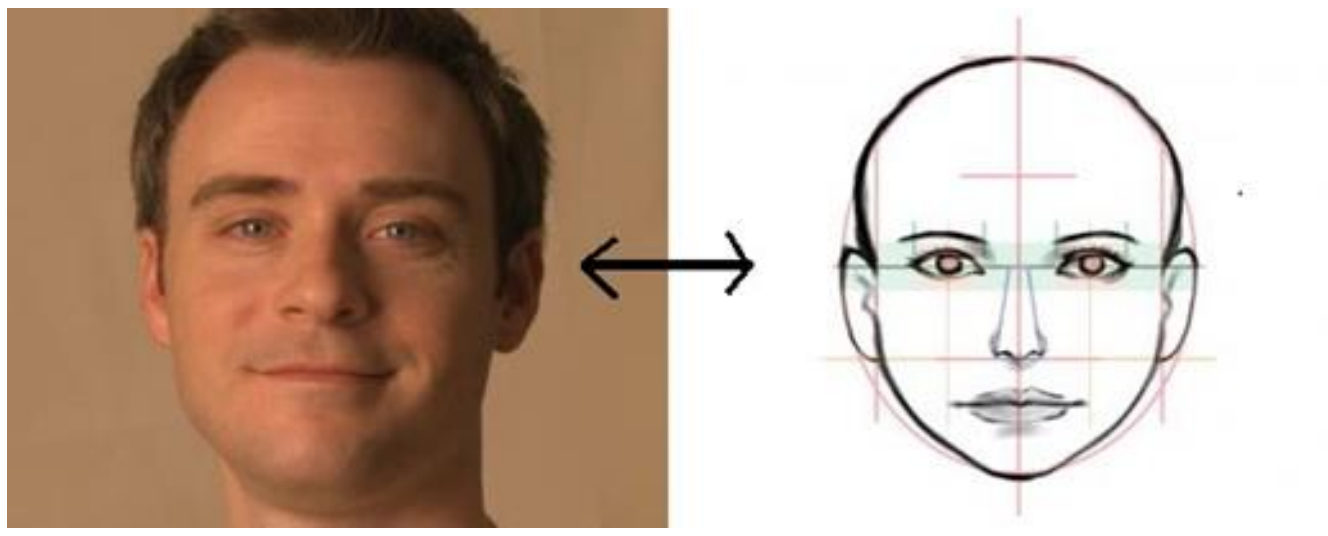

Figure 5. Pictorial representation of the set of human faces reduced to the extracted feature sets of atomic symbols.

Figure 5 is a pictorial representation of the set of human faces reduced to the extracted feature sets of atomic symbols [i.e. Ear- Eye- Nose- Eye- Ear]. In the class of Human, various unit elements bear different set of frameworks within the same alignment which helps in distinguishing the elements.

*a. Noisy Data in Parsing and Recognition*

Differing from the fundamental belief gathered from the earlier part of this paper, the SP system can also handle sequence of video images for detection and tracking.

In Fig. 4, we have shown that the SP System is quite adaptive to detect and omit errors, such as, partial occlusion, noisy data handling etc. As briefly illustrated in (Fig. 6), the newly formed pattern on the arrival of new raw data line in row 0 remains the same as in (Fig. 2) instead of the replacement of 'm' for 'n' in 'k i t t e n s', the absence of the 'w' in 't w o' and within the word 'p l a y', the erroneous addition of 'x'. In spite of these errors and noise addition, SP62 model derives the best possible multiple alignment, as shown in (Fig. 6), which in turn reflects the correct initial alignment of the feature set.

*b. Family Resemblance*

An alternative idea is, SP systems strongly accommodates 'Family resemblance', in terms of polythetic concepts: the method of parsing the raw data for visual detection and recognition is not dependent on the presence of any key feature or combination of features, as well as in the absence of it [15], [33].The system is well susceptible to errors in form of partial occlusion, noisy data etc., via searching for its optimal solutions [Sections III.A (a)], as it partly allows for the requirement of knowledge based alignments that may have various alternatives at any given point within the structure. Most of the SP system frameworks are polythetic. Although possession of a pair of legs seems to be a key feature to identify the concept of 'Human', yet the system should recognize Sam as a Human, even with partial occlusion which visually depicts a loss of one Leg. Similarly, this method is adapted for most of the concepts in any visual systems. In any logical system that aims to achieve human-like vision, the concept of 'Family resemblance' or polythetic is very essential.

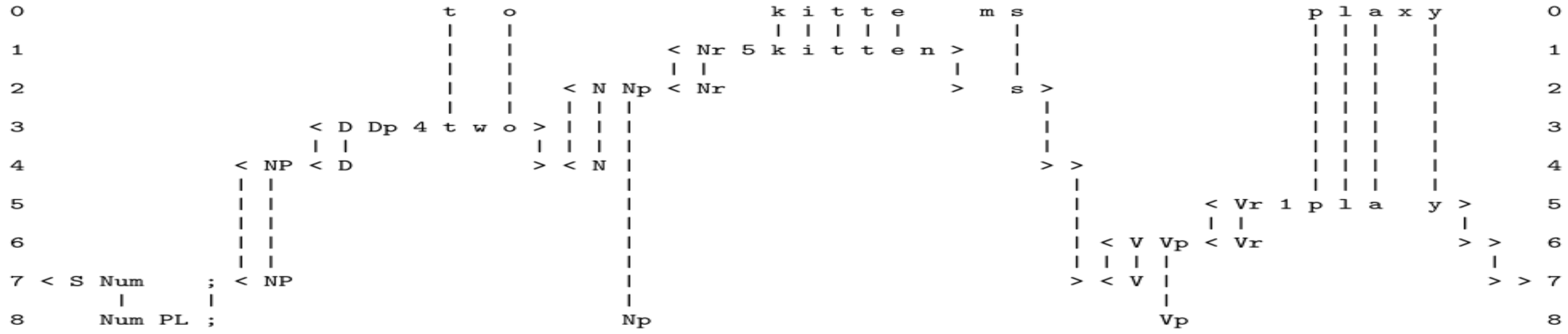

Figure 6. Noisy Data in Parsing and Recognition.
Best possible multiple alignment is extracted despite of presence of noise. [26]

### *c. Hierarchies and their integration*

SP systems consists of various multiple alignments representing various objects of interest and domain of interests, which is simple yet effective for object detection and recognition in any visual system. Representation of classes of objects and processing of class hierarchies, part-whole hierarchies and their integration, as mentioned by [15], [26]. In Figure 7 (a, b, c), a multiple alignment of all the parts and sub parts of a human body is shown. This does not illustrate the visual appearance of a human body but it is sufficient to represent and process all the relative information to form a human body out of it.

It is safe to assume that, this system being efficient to work with two dimesons, has the capability to process all the parts and sub parts and relating to a hierarchy based on information extracted from raw data in form of multiple alignment. The integrated form is shown in [Fig 8].

### *d. Scene Analysis*

Scene analysis is broadly a primary subsection of knowledge parsing, for example: In the process of analyzing a sea beach, high-level feature alignments are recorded of things that may typically be seen in a sea beach (i.e., rocks, boats, sea, beach, sky and so on), with more logical detailing about its existence in the raw data. The complications we face, as suggested by [35] in the process of a scene analysis are:

- Partial occlusion is one of the primary anomalies in the process of scene analysis. In a typical sea beach, various feature points of the data can be partially obscured by other mutually exclusive features, creates an ambiguity about the domain of the scene, i.e. a boat is partially occluded by other features, such as, waves, sea birds etc.
- Variability of the locations in the scene of all the feature points creates ambiguity about the scene, i.e. A boat may be on the beach or in the sea.

Although, people relate to the aforementioned anomalies quite easily, but in complicated scenarios, 'naïve' kind of parsing systems fail to address such issues. The SP systems retains these aspects, carefully, with respect to scene analysis in following ways:

- We have already established in Section II [A (a)], SP systems are well equipped with handling errors, noise, omissions, commissions and substitutions. Thus, it is safe to assume that, the SP models that are comprehensive to work with patterns in two dimensions, can handle partial visibility of the objects and recognize them successfully in the subsequent frames.
- Inconsistency of scenes captured from any real-time video is similar to parsed sentences in natural language. SP systems, among other artificial systems, is capable of supporting the system with relevant information about the scene in form of wide variety of multiple alignments and phrases containing recursive forms. This principle is well applicable to Vision related domains. For example: "Politics is the art of looking for problem, finding it everywhere diagnosing it incorrectly and applying the wrong remedies."
- Existing knowledge is not always palpable to varying domains and raw data line. In such scenarios, the system may learn from its experiences, as briefly mentioned in Section III (B).

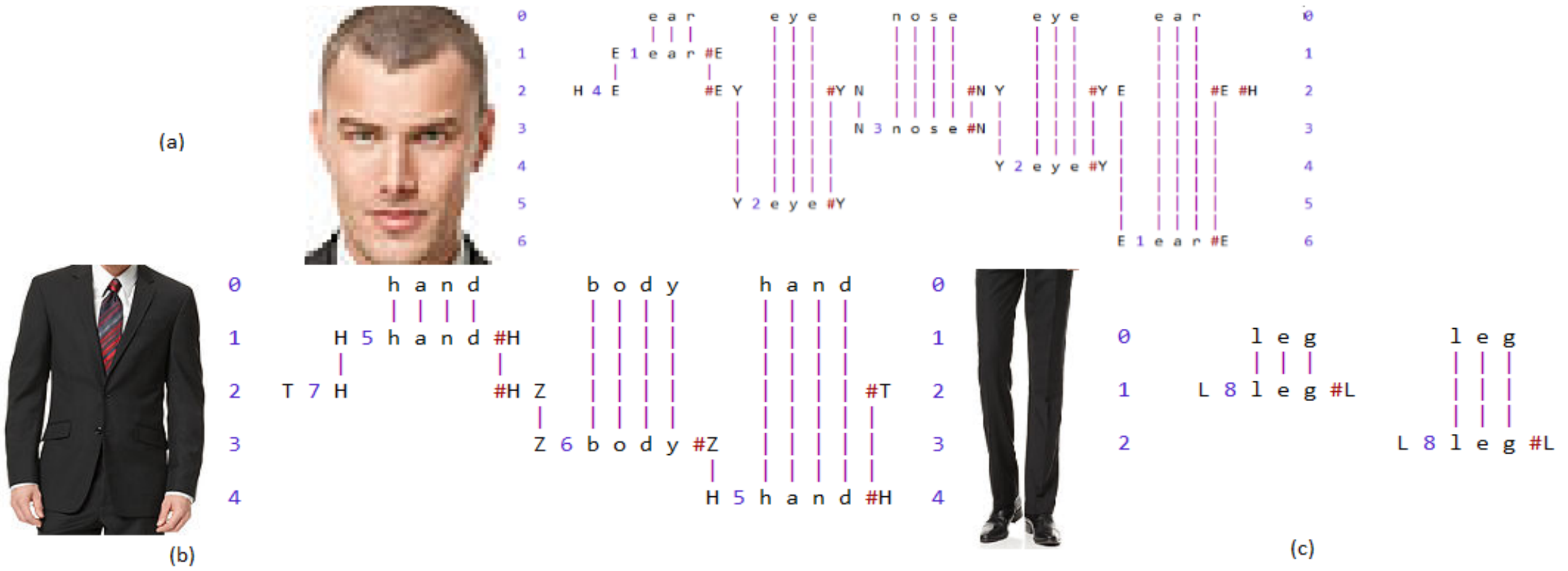

Figure 7. (a): Multiple alignment of human head.
(b): Multiple alignment of human torso.
(c): Multiple alignment of human leg.

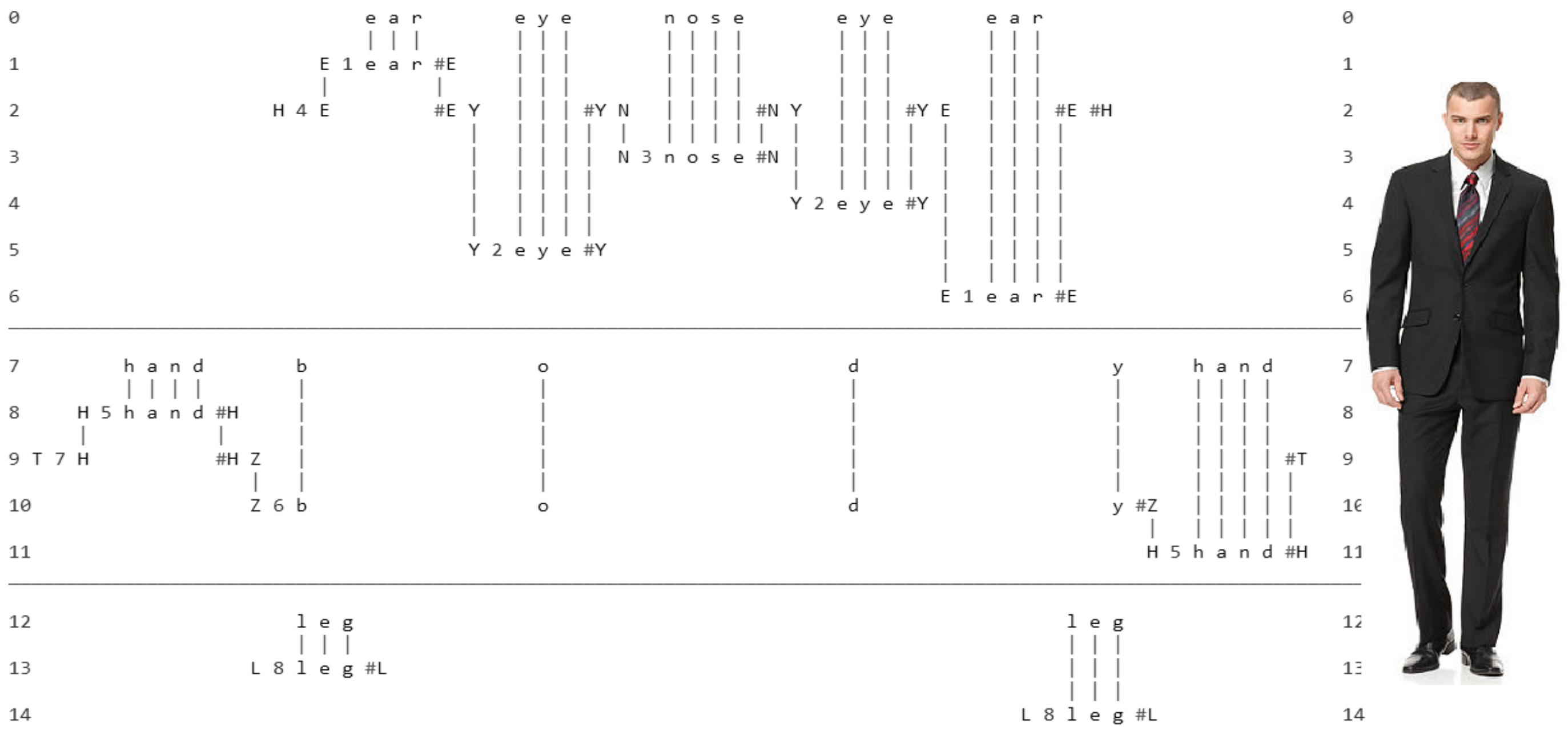

Figure 8. Integrated form of Fig. (7) Human body along with its multiple alignment structure.

*B. Unsupervised learning and 'DONSVIC' principle*

Learning is an essential part of computer vision since gaining new information and to monitor the changes happening around the world are primarily done by vision. In general, it is quite evident that, learning through vision is mostly unguided in nature that is 'unsupervised'. Intervention of a 'teacher' is not required when it learns through vision. The classification of samples from simpler to complex ones and provision of 'negative' samples are not required. We try and get information through our vision and try to comprehend that in our knowledge to make sense of it in the best possible way.

Unsupervised learning has been developed in the SP framework and rightly so, it works better than most of the well-developed knowledge based frameworks. In this section, we would like to demonstrate how unsupervised learning is developed in SP framework and applied in the vision via the 'DONSVIC' principle of unsupervised learning.

While dealing with our surroundings, there are certain kind of structures or objects or class of objects, that appear more useful and prominent than the others: for better understanding of visual appearances of 'discreet' objects (i.e. 'person', 'tree', 'house' etc.). These 'natural' kind of structures or class of objects are substantial in our information processing and compression of sensory information, which in turn, provides the key to learn and discover new objects. Even though, popular LZW algorithms based on information compression from JPEG images are more reliant to recognize words or objects in form of information and interpreting the knowledge in the application domain, but they are mostly designed to work on low-powered machines.

In SP systems, programs are slower yet thorough and reveals natural structures in detail, as briefly explained below:

- Parsing of a corpus of natural language text, unsegmented, created by the MK10 program (Wolff, 1977), using only the information provided by the corpus of natural language text without any supervised knowledge provided dictionary or knowledge base about the structure of the language (Fig. 9). Even without all of its punctuations and spaces separating words are removed from the corpus, the system works exceedingly well in revealing the word structure of the text.
- Similarly, the SP system works perfectly well, significantly better than chance, in detecting phrase structures from a corpus of natural language texts without reasonable punctuations or spaces, but with a symbol replacing words for its grammatical category. The process of replacing is done by a trained linguistic analysis, but the discovery of the structures of new phrases is done by the system, without supervision.
- Derivation of a plausible grammar, from an unsegmented corpus of artificial language without any assistance is done by The SNPR program. The SNPR program for grammar discovery can learn new words from the text corpus, grammatical categories and the structure of phrases and words.

MK10 [26] and SNPR [26] programs are designed and equipped to search through the variety of alternatives among patterns which may be unified and matched to retain the set of patterns that yield a higher level of compression. This principle is not only applicable to discovery of words, grammars and pattern of words from artificial languages, but also in the area of vision: discovery of objects in images, class of entity in various kind of data. Principle is broadly termed as 'the discovery of natural structures via information compression', or 'DONSVIC'

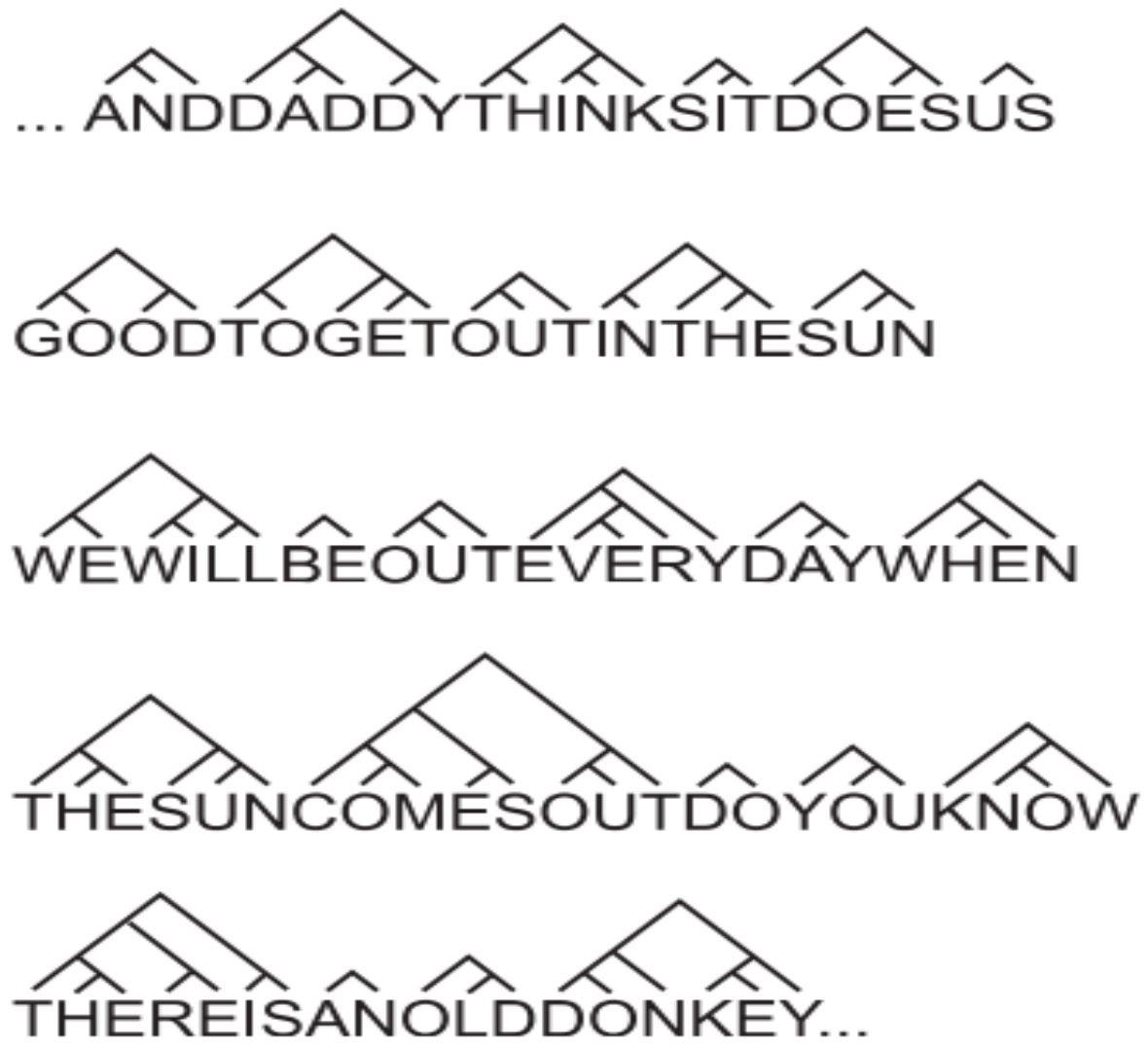

Figure 9. Part of a parsing created by program MK10. [26]

A radically new conceptual information compression framework is developed with the concept of multiple alignment. As mentioned earlier, the SP70 system works on multiple alignments, deriving Old patterns from corpus of natural language texts and comprehending them into the knowledge base to create New Old patterns with economical and exceptional low-scoring tests.

```
0      t h a t g i r l r u n s      0
       | | | |         | | | |
1 A 1  t h a t b o y   r u n s  #A 1

(a)

B 2 t h a t #B
C 3 b o y #C
C 4 g i r l #C
D 5 r u n s #D
E 6 B #B C #C D #D #E

(b)
```

Figure 10. (a): Simple Old pattern is derived from SP70.
(b): Newly formed Old pattern. [26]

SP learning system is illustrated schematically in Fig. (10). The SP system, as an abstract system works like a human brain, receiving 'new' information via its senses and deriving Old patterns in form of information. Suppose, the system hears someone saying "t h a t b o y r u n s". If the system never heard anything similar, then it stores New information as a relatively straightforward copy, as shown in row 1 of the multiple alignment in Fig. 10.

### *C. Tracking with Particle Swarm Optimization (PSO)*

The PSO framework provides an effective way to track multiple object that are detected and recognized from aforementioned method (SP artificial systems). First, for singular object tracking, following analogies need to be assumed:

- The groundtruth of an object and surrounding region can be considered as ecological properties.
- State space particles correspond to a particular species.
- Each particle's observation likelihood and fitness capability of a particular species is analogous.

For multiple object tracking, these postulates can be easily extended by creating a tracker for each object. These trackers are managed independently. In case of occlusion, support regions of concerning objects may overlap, which implies, the intersectional area between two species are elementary to both. Subsequently, the repulsion and competition among the species arise as both of them aspire to the same resource, the stronger one has higher probability of winning the competition.

During the course of video scene there may be overlap between two object areas due to occlusion, and the related features between them become ambiguous. To handle this complication, we design a multiple-species-based PSO algorithm as suggested by [19]. The principle idea behind this approach is to divide the groundtruth particles of the object into various species according to the species object numbers and successfully model the relations and the partial visibility among varied species. Detailed description of the species inspired PSO algorithm is briefly described in the following sections.

#### *a. Problem Construction:*

Let us consider, M number of objects, surrounded with N number of particles, constitute a set $\chi = \left\{ x_{t,k}^{i,n}, i = 1, \ldots, N, k = 1, \ldots, M \right\}, \boldsymbol{O} = \left\{ o_{t,k}^{i,n}, i = 1, \ldots, N, k = 1, \ldots, M \right\}$. Here $t$ is the 2-D translation parameter. Formula of multiple object tracking is as follows:

$$\chi^* = \arg\max_i p(\boldsymbol{O}|\boldsymbol{\chi}) \tag{1}$$

By independently maximization of the individual observation likelihood, the above optimization may be simplified, in case of no occlusion.

In case of no occlusion, the above optimization may be simplified by maximizing the individual observation likelihood independently (here, we drop the superscript i, n for simplicity):

$$x_{t,k}^* = \arg\max_{x_{t,k}} p\left(o_{t,k} \middle| x_{t,k}\right), k = 1, \ldots, M \tag{2}$$

#### *b. Competition Model:*

When different object obscure one another, there is an overlap between corresponding support regions. In these circumstances, the competition between two objects elevates to subjugate the overlapping part (Fig. 11). In order to effectively design the competition phenomenon, the visual problem needs to be merged with the competition process.

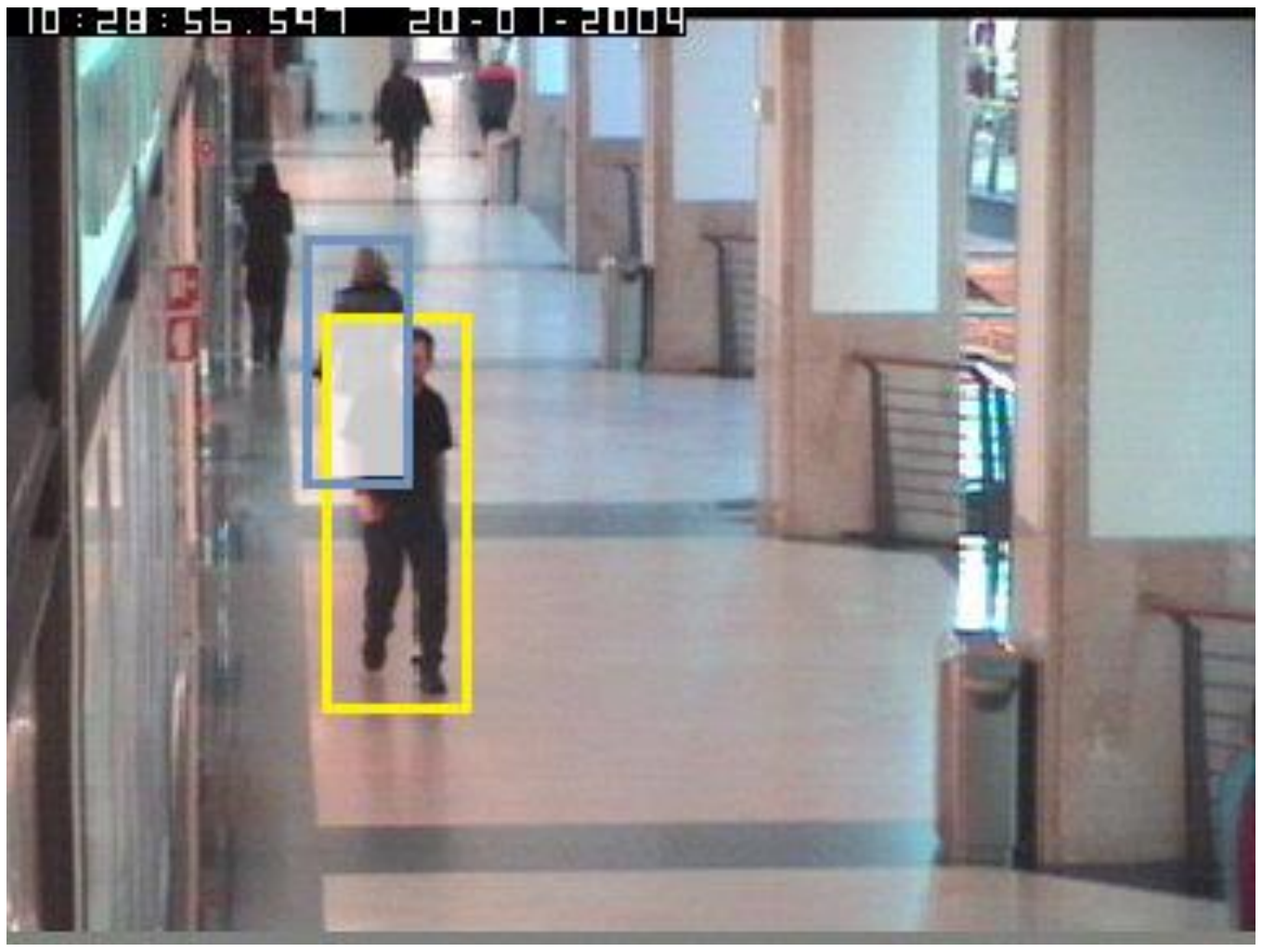

Figure 11. Overlap between two object areas.

To evaluate the fitness value on the overlapping part as the competition ability, the overlapping part is viewed as a whole and projected onto the learned subspace corresponding to each object. We define the power of each object or species in following manner:

$$power^{k} = p\left(\hat{o}_{t,k}\middle|x_{t,k}\right) = exp\left(-\left\|\hat{o}_{t,k} - \widehat{U}_k \widehat{U}_k^T \hat{o}_{t,k}\right\|^2\right), \quad (3)$$

where $k$ and $\widehat{U}_k$ are the overlapping part of the object and its corresponding subspace respectively. In a similar way, the interactive likelihood of object $k_1$ over the overlapping regions can be calculated:

$$\underbrace{p\left(\hat{o}_{t,k_1}\middle|x_{t,k_1}, x_{t,k_2}\right)}_{interactive\ likelihood} = \frac{power^{k_1}}{\sum_{i=1,2} power^{k_i}}. \quad (4)$$

The mutual likelihood of each species describes the competition ability. Higher the competition ability of a species more like it is to win the competition. It means that the species which won the competition is more likely to be of the object that was occluding the other object species involved.

*c. Annealed Gaussian Based PSO (AGPSO)*

An annealed Gaussian based PSO algorithm [21] is considered in this paper, as in conventional PSO requires careful and fine tuning of various parameters. In this algorithm, the particles and corresponding velocities are updated as stated below:

$$v^{i,n+1} = |r_1|(p^i - x^{i,n}) + |r_2|(g - x^{i,n}) + \epsilon \quad (5)$$

$$x^{i,n+i} = x^{i,n} + v^{i,n+1} \quad (6)$$

where $|r_1|$ and $|r_2|$ being the absolute values of the samples from Gaussian probability distribution N(0, 1). This is zero-mean Gaussian disturbance that stops the algorithm from getting trapped in local optima. With the help of adaptive simulated annealing, the covariance matrix of $\epsilon$ is changed [34]:

$$\textstyle\sum_{\epsilon} = \sum e^{-cn} \quad (7)$$

Here, a transition distribution is predefined, and Σ is its covariance matrix, annealing constant *c*, and iteration number *n*. The components in Σ decrease in proportion to the iteration number which results in a fast rate of convergence. When $k_1$ and $k_2$ occlude each other at time t, a repulsion force is added to the evolution process of particles, and subsequently the iteration step for $k_1$ becomes as follows:

$$v_{t,k_1}^{i,n+1} = |r_1|\left(p_{t,k_1}^{i} - x_{t,k_1}^{i,n}\right) + |r_2|\left(g_{t,k_1} - x_{t,k_1}^{i,n}\right) + |r_3|F_{\overrightarrow{k_2,k_1}} + \epsilon \quad (8)$$

$$x_{t,k_1}^{i,n+1} = x_{t,k_1}^{i,n} + v_{t,k_1}^{i,n+1} \quad (9)$$

where the parameter $r_3$ is Gaussian random number sampling from N (0, 1). The third term on the right-hand side of the above equation depicts the shared effect between object $k_2$ and $k_1$. In other words, the competition phenomenon on the observation level has been modelled in this paper. Also, the competition model of state space has been modelled to drive the evolution process of the species in the right direction.

*d. Updating of the Appearance Model Selectively*

In most of the tracking algorithms [24], [29], appearance models are not updated during occlusion. However, the appearance of the object under occlusion may change, and that can cause the tracker to fail to recapture the object appearance if it is not occluded anymore. A selective updating algorithm is implemented to cope with the appearance changes during occlusion: 1) pixels belonging to the visual part of the objects are cumulatively updated in the normal way, 2) pixels that are part of the overlapping region (Fig. 11) are projected onto the subsequent subspace of each object. Then the errors due to the reconstruction are calculated. If this error is smaller than a predefined threshold for pixels inside the overlapping area, then it is again updated in the subsequent subspace.

Due to this careful modelling of the updating strategy, the appearance changes can be easily accommodated, allowing more persistent tracking throughout the video stream.

## IV. Experimental Data and Analysis

Proposed method is observed on benchmark datasets to demonstrate the robustness and adaptivity of SP systems to cope with the varied challenges offered by them. Brief description of the benchmark datasets is shown followed by the experimental settings and experimental observations. The SP framework is implemented in SP70 Virtual Machine, provided by Wolf. J. G for research purposes *(http://www.cognitionresearch.org/sp.htm#SOURCE-CODE)* on VC++ with an Intel core 5th Gen i7, 2.10 GHz processor with 6 Gigabytes of RAM and 2 Gigabytes NVDIA GeForce GPU.

### A. Observational Datasets

We observed the efficiency and robustness of the proposed algorithm in natural vision on benchmark datasets on some of the TB100 sequences, namely, Walking2 [29], Jogging (1,2) [22] and David [26]. In the following section, we discuss the aforementioned datasets and the challenges the datasets possesses in natural vision; especially object detection, recognition and tracking.

Walking2 [29] possess challenges like: Scale Variation (SV), Occlusion (OCC) and Low Resolution (LR). The video contains 500 frames, each of dimension 384x288 pixels. The video is of some people walking down the corridor of an office interior.

The David dataset [26] is much more challenging compared to Walking2 [29], as one has to consider in-frame challenges, like: Illumination Variation, Scale Variation, Occlusion, Deformation (DEF), Motion Blur (MB), In Plane Rotation (IPR) and Out Plane Rotation (OPR). The dataset contains 770 video frames of 320x240 resolution. The video is of a person walking down the corridor of an office interior.

Jogging (1,2) [22] contains complications like Deformation, Occlusion and Out Plane Rotation. The video contains 307 frames, each of dimension 352x288 pixels. The video is of two pedestrian jogging down the road.

In the following section, we demonstrate the adaptability and efficiency of SP theory to cope with the challenges provided by these datasets and how accurately problems can be dealt with.

### B. Experimental Observation and Analysis

We observe our method on the aforementioned datasets with the SP70 Virtual Framework and others, as mentioned above. Initially, we have observed the original videoframes and the challenges it possesses. Firstly, we train the system with the multiple alignments of our object of interest, i.e. Human, as described in Fig. (8). The system stores the information in form of Old pattern, that is subjected to detect and to be tracked in the test domain. A pictorial demonstration is shown in Fig. (12).

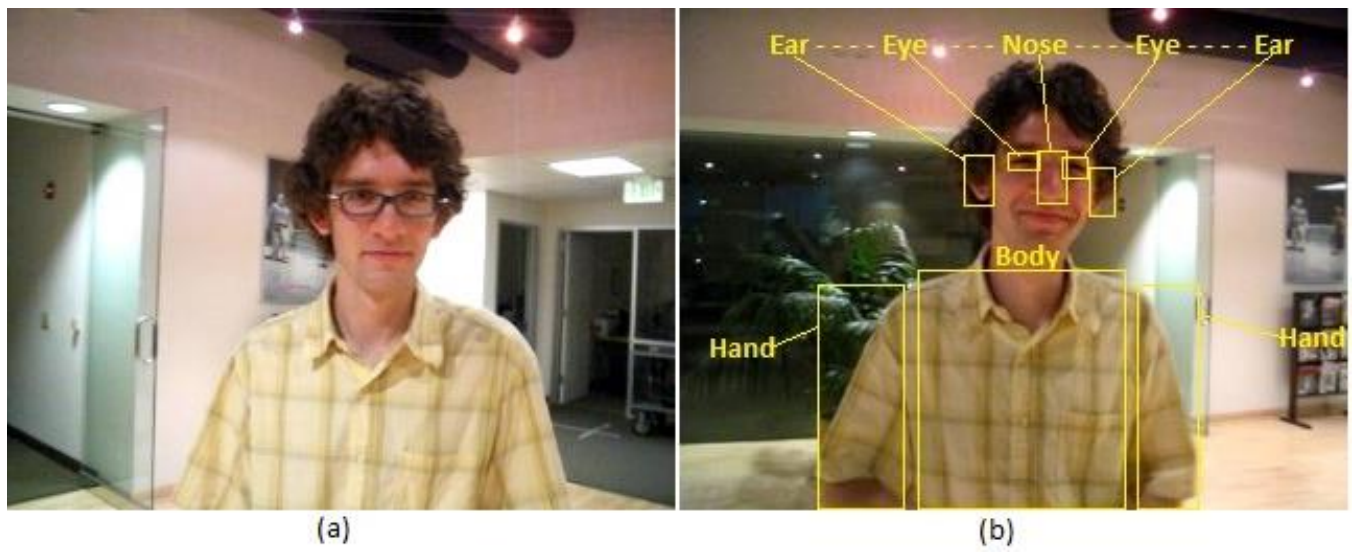

Figure 12: Atomic symbol representation of different body parts.

As mentioned earlier, SP systems do not illustrate the visual appearance of human body, but it processes the visual inputs provided from other systems with the help of multiple alignment concept and can successfully interpret different body parts to gather the information associated with the human body (Fig. 13). Thus, in turns, helps to string together the part-whole hierarchies and class hierarchies of different parts of human body.

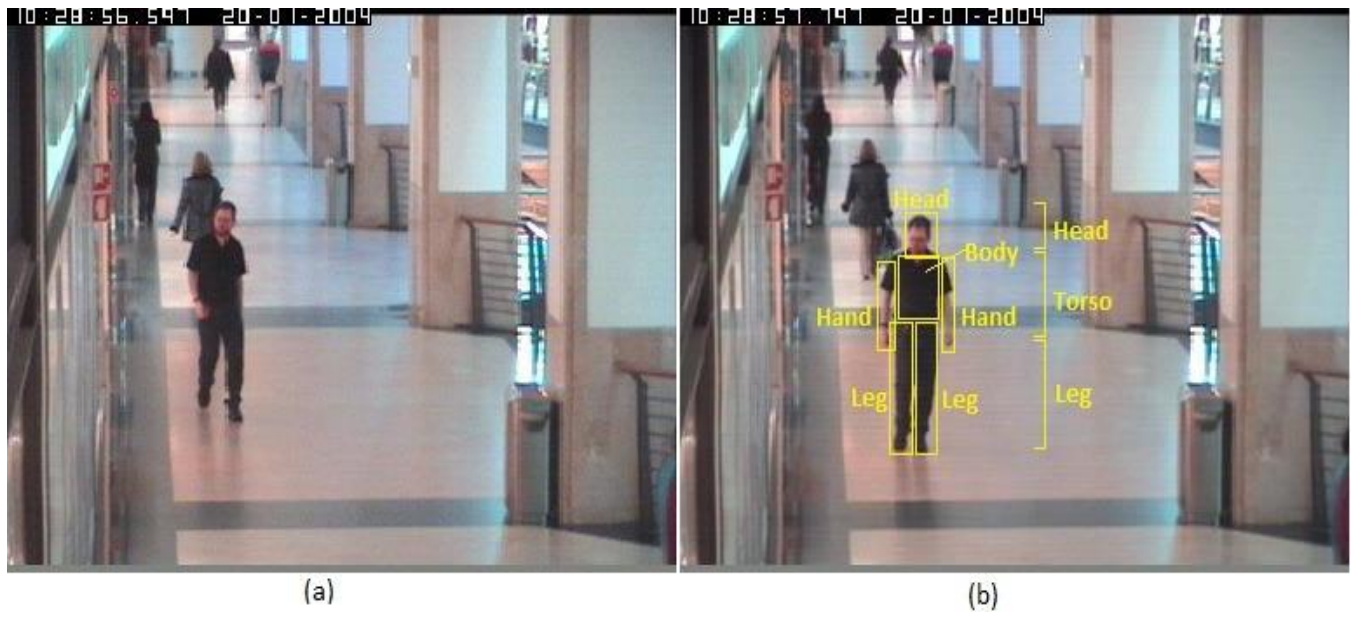

Figure 13: Full body symbolic representation.

In a video scene, our object of interest may be occluded by some other object native to that domain. In these cases, SP system matches the multiple alignment stored as Old information with the visible parts from the target objects and derives the remaining part of object despite its invisibility due to occlusion. As mentioned earlier, in the Jogging dataset two pedestrians are being tracked. The primary challenge we have faced in this particular video sequence, is that the pedestrians are partially occluded by a post in some of the video frames (Fig. 14). In such scenarios, the SP system processes the visible parts of the object of interest (i.e. partial head, torso, leg), and matches them with the previously stored information in the Old pattern to derive the supposedly New pattern in order to make sense of the current knowledge base.

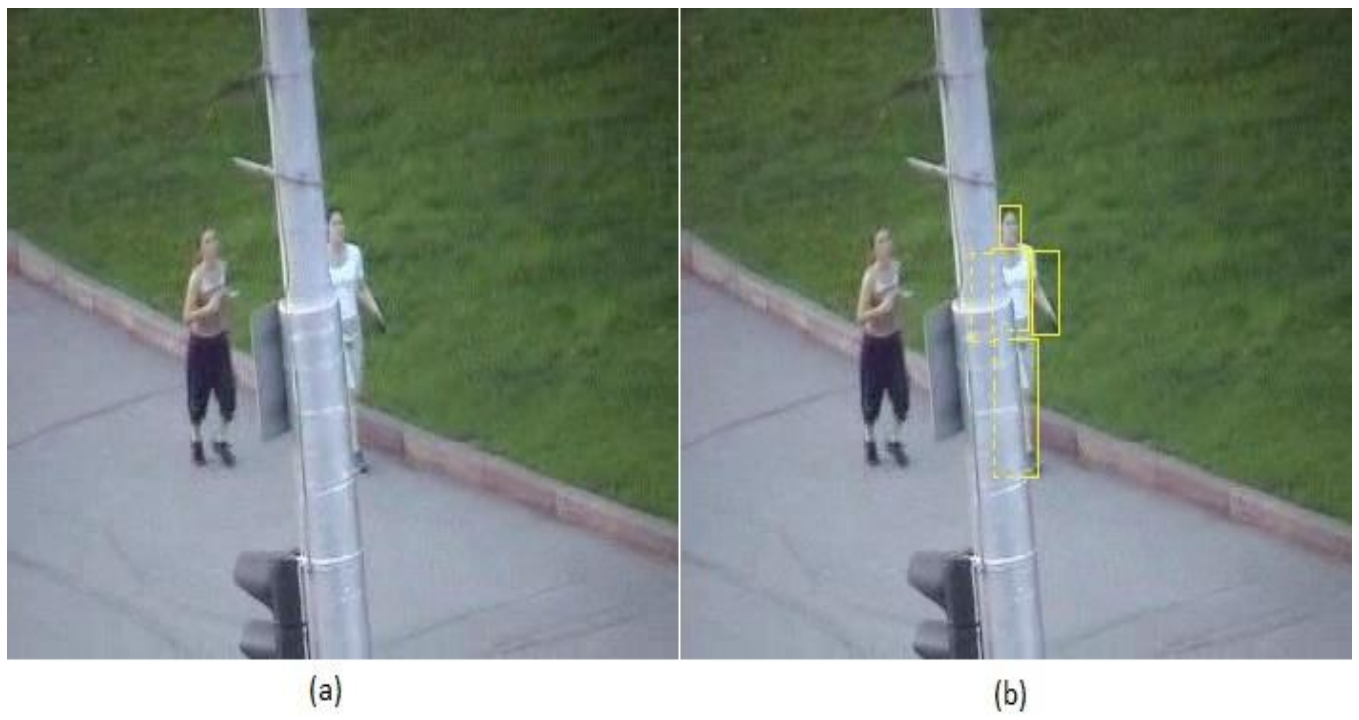

Figure 14: Occlusion handling.

The multiple alignment structure for partial occlusion, as mentioned above, is derived as Fig. 15. Consequently, the extracted alignment of the object of interest is tracked via species inspired Particle Swarm Optimization (PSO) in successive frames.

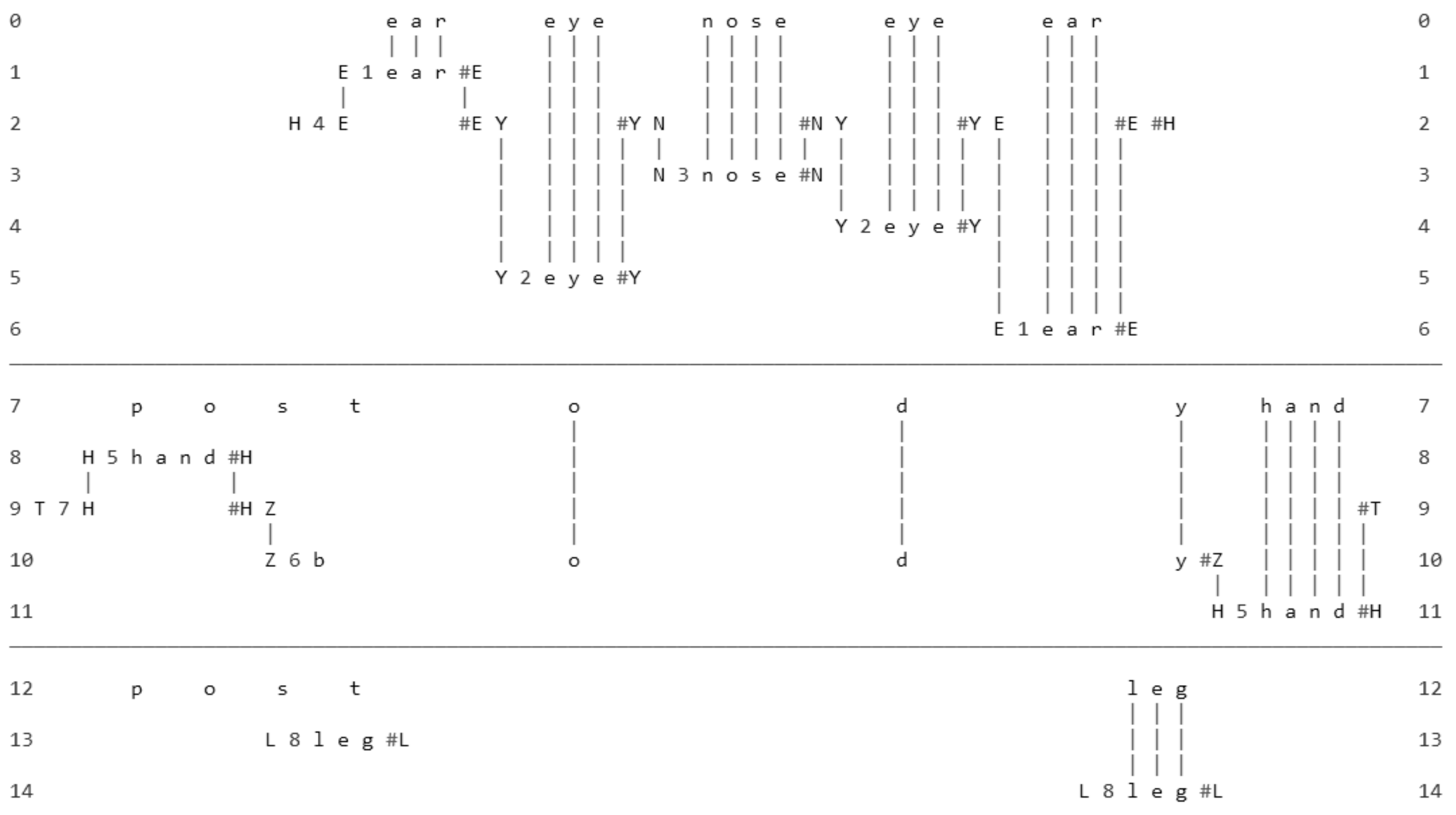

Figure 15: Best possible multiple alignment of a Human body is extracted despite of presence of occlusion.

We have compared the performance of the present algorithm our vision system with the existing state-of-the-art algorithms and the result of comparison is briefly shown in Table 1. Specific parameter construction and technical details are incorporated while implementing various algorithms, hence, we have chosen only those algorithms whose source code or binary code was available publicly.

For comparisons, various parameters of considered trackers are fixed for all the experiments. The evaluation result of our experiment represents the lower bound of the average tracking performances. Despite of having varied approaches followed by any individual tracker, the primary features are somewhat similar and shared by all of state-of-the-art methods.

Table 7 represents the average number of failures and average number of success of our algorithms with respect to six algorithms ([12], [13], [35], [20], [7], [9]). For handling occlusion, the overlap threshold of 0.65 with respect to all it attributes is considered. Primarily, the initial runs from the beginning are generated at 60 frames, whereas, the virtual runs are raised for our threshold and experiment window. The trackers' performances are steady over large variations of experiment window by keeping a window size of 90.

As the overlap threshold increases, the number of failures increases and the success rate increases at first and then decreases rapidly over the couple of successive frames. To When the overlap threshold is low, a tracker is not restarted even when it actually loses track of a target object and the model of a tracker is likely to be incorrect.

The average success rate increases when the threshold values are low, since re-initialization significantly helps in keeping track of our object of interest even after failures of a frame or two.

Our algorithms perform better over other state-of the-art algorithms in various benchmark datasets. To support our claim, we have tested our tracking algorithm against the existing ones and formulated the quantification in Table 2 (David), 3 (Walking2), 4 (FaceOcc1), 5 (Jogging). In all the cases, our algorithm outperforms the existing state-of-the-art trackers.

Our method can steadily deal with partial occlusions and deduce the partial relationship of that object with its domain and surrounding based on its Old information and alignments from its knowledge base and interactive likelihood of species. The results, as demonstrated in Figure 11 and evaluated in Table 7, support our claim. The singularity of the object with relatively greater fitness value on the overlapping part will occlude the other object, more likely.

A brief illustration of other advantages presented in this paper, a quantitative analysis with other state-of-the-art methods is presented in Table 8: average number of failures in 1000 frames under occlusion, RMSE (root mean square error) of Position i.e. the distance between the estimated position and the groundtruth. It is quite evident from the results that our algorithm achieves optimum accuracy in localizing the object of interest, performs better under occlusion and shows significant breakthrough against the existing state-of-the-art tracking algorithms.

Table 1. Evaluated Tracking Algorithms

| Trackers | Representation | | | | | | | | | | Code | | | |
|---|---|---|---|---|---|---|---|---|---|---|---|---|---|---|
| | Local | Template | Color | Histogram | Subspace | Sparse | Binary or Haar | Discriminative | Generative | Model Update | C/C++ | MATLAB | FPS | Published |
| ASLA | ✓ | ✓ | | | ✓ | ✓ | | | ✓ | ✓ | ✓ | ✓ | 8.5 | ‘12 |
| BSBT | | | | | | | H | ✓ | | ✓ | ✓ | | 7.0 | ‘09 |
| CXT | | | | | | | B | ✓ | | ✓ | ✓ | | 15.3 | ‘11 |
| DFT | ✓ | ✓ | | | | | | | ✓ | ✓ | | ✓ | 13.2 | ‘12 |
| IVT | | ✓ | | | ✓ | | | | ✓ | ✓ | ✓ | ✓ | 33.4 | ‘08 |
| LIAPG | | ✓ | | | ✓ | ✓ | | | ✓ | ✓ | ✓ | ✓ | 2.0 | ‘12 |
| LOT | ✓ | | ✓ | | | | | | ✓ | ✓ | | ✓ | 0.7 | ‘12 |
| LSHT | ✓ | | ✓ | ✓ | | | H | | ✓ | ✓ | ✓ | | 20 | ‘13 |
| LSK | ✓ | ✓ | | | ✓ | ✓ | | | ✓ | | | ✓ | 5.5 | ‘11 |
| LSS | ✓ | ✓ | | | ✓ | ✓ | | | ✓ | ✓ | | ✓ | 15 | ‘13 |
| MIL | | | | | | | H | ✓ | | ✓ | ✓ | | 38.1 | ‘09 |
| MTT | | ✓ | | | ✓ | ✓ | | | ✓ | ✓ | ✓ | | 1.0 | ‘12 |
| ORIA | | ✓ | | | ✓ | | H | ✓ | | ✓ | ✓ | | 20.2 | ‘11 |
| PCOM | | ✓ | | | ✓ | ✓ | | | ✓ | ✓ | | ✓ | 20 | ‘14 |
| SMS | | | ✓ | ✓ | | | | | ✓ | | ✓ | | 19.2 | ‘03 |
| Proposed Method | | | ✓ | ✓ | | | | ✓ | | ✓ | | ✓ | **34.3** | – |

Table 2. Tracking Accuracy on David Dataset

| **Approach** | **Year** | **Accuracy** |
|---|---|---|
| ASLA [12] | 2012 | 82.24% |
| DFT [13] | 2012 | 81.31% |
| IVT [35] | 2008 | 83.86% |
| MIL [20] | 2009 | 86.19% |
| PCOM [7] | 2014 | 78.37% |
| LSS [9] | 2013 | 79.34% |
| **Proposed method** | | **91.3%** |

Table 3. Tracking Accuracy on Walking2 Dataset

| **Approach** | **Year** | **Accuracy** |
|---|---|---|
| ASLA [12] | 2012 | 88.2% |
| DFT [13] | 2012 | 87.9% |
| IVT [35] | 2008 | 89.8% |
| MIL [20] | 2009 | 90.2% |
| PCOM [7] | 2014 | 88.3% |
| LSS [9] | 2013 | 90.7% |
| **Proposed method** | | **94.7%** |

Table 4. Tracking Accuracy on FaceOcc1 Dataset

| **Approach** | **Year** | **Accuracy** |
|---|---|---|
| ASLA [12] | 2012 | 91.3% |
| DFT [13] | 2012 | 88.3% |
| IVT [35] | 2008 | 93.9% |
| MIL [20] | 2009 | 95.1% |
| PCOM [7] | 2014 | 89.4% |
| LSS [9] | 2013 | 87.2% |
| **Proposed method** | | **97.7%** |

Table 5. Tracking Accuracy on Jogging Dataset

| **Approach** | **Year** | **Accuracy** |
|---|---|---|
| ASLA [12] | 2012 | 85.3% |
| DFT [13] | 2012 | 86.8% |
| IVT [35] | 2008 | 90.7% |
| MIL [20] | 2009 | 92.4% |
| PCOM [7] | 2014 | 86.7% |
| LSS [9] | 2013 | 88.5% |
| **Proposed method** | | **93.7%** |

Table 6. Detection Accuracy on Dudek Dataset

| **Approach** | **Year** | **Accuracy** |
|---|---|---|
| ASLA [12] | 2012 | 86.7% |
| DFT [13] | 2012 | 86.3% |
| IVT [35] | 2008 | 88.9% |
| MIL [20] | 2009 | 91.2% |
| PCOM [7] | 2014 | 86.7% |
| LSS [9] | 2013 | 87.2% |
| **Proposed method** | | **92.8%** |

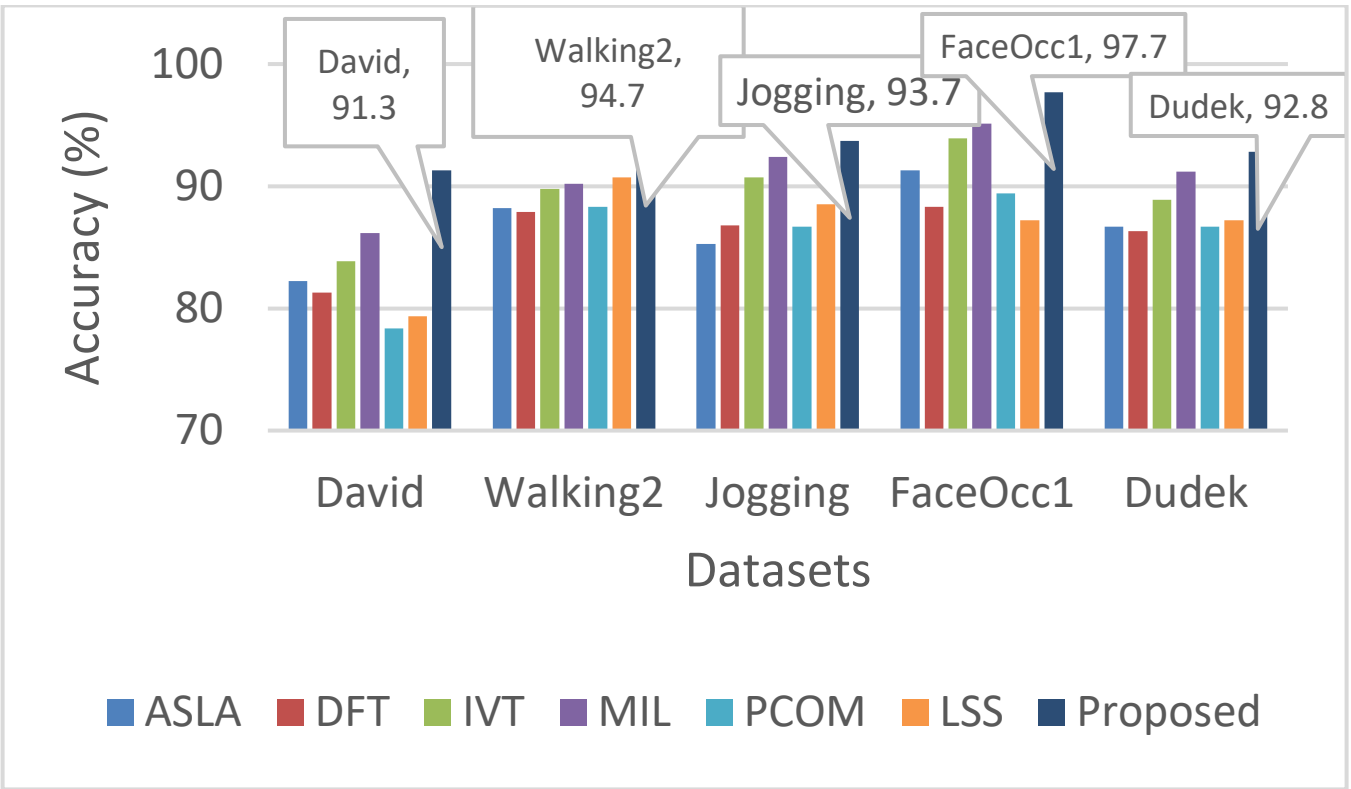

Figure 16. Accuracy analysis against various datasets.

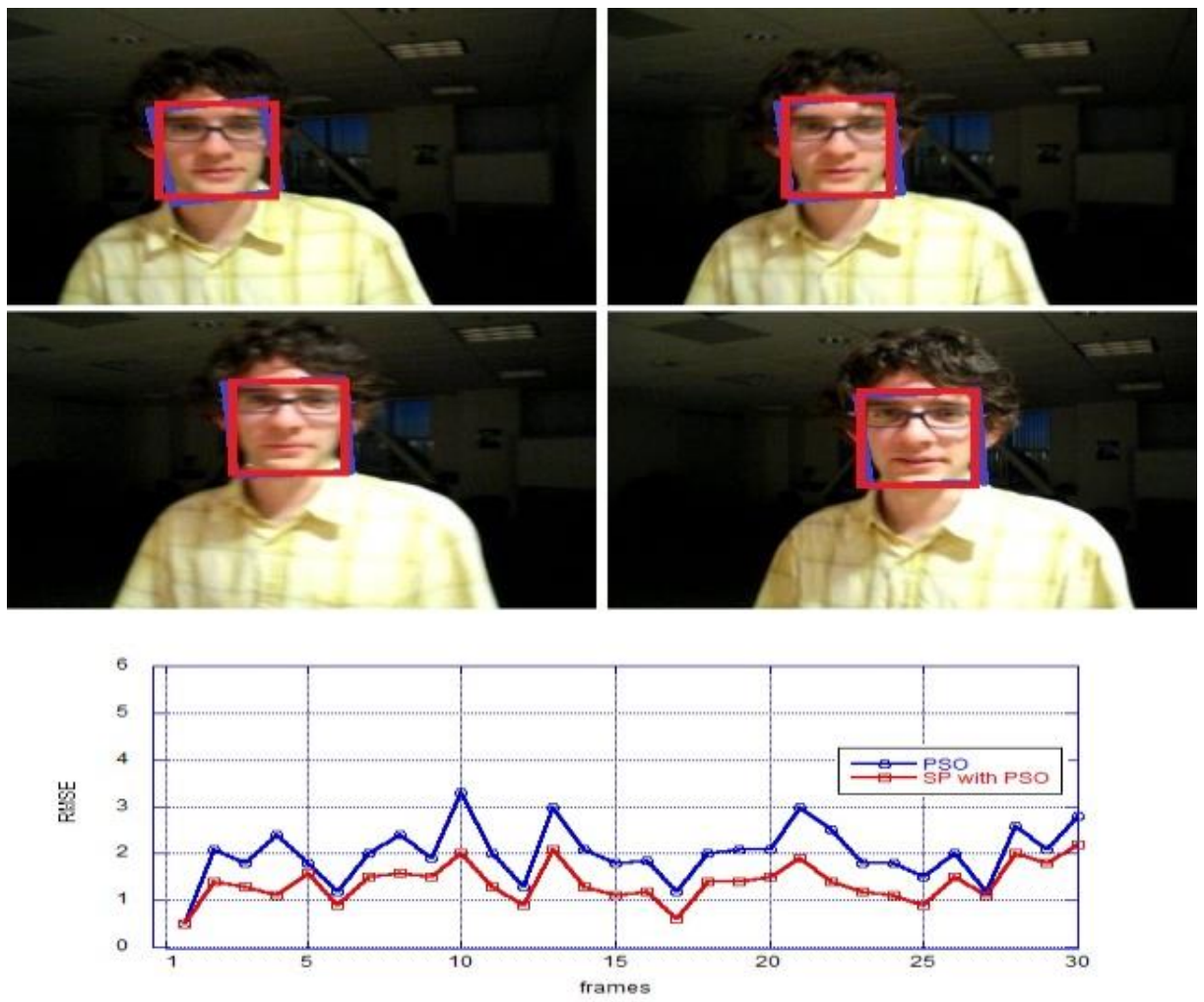

Figure 17. Tracking performance on David Dataset and comparison with respect to RMSE.

Figure 16 demonstrates the tracking accuracy of aforementioned algorithms in various benchmark datasets. The tracking performances of traditional PSO and species inspired PSO with SP theory on a fast-moving face is graphically represented in Figure 17. Here we used RMSE (root mean square error) as the performance metric, from which we can see that the latter can achieve a remarkably higher tracking accuracy due to the amalgamation of SP theory of intelligence and species inspired PSO.

Table 7. Performance of different algorithms on different attributes

| Trackers / Features | ASLA [12] | DFT [13] | IVT [35] | MIL [20] | PCOM [7] | LSS [9] | Proposed method |
|---|---|---|---|---|---|---|---|
| Scale Variation (SV) | 54.0 / 3.9 | 47.9 / 5.9 | 47.1 / 5.3 | 44.5 / 6.5 | 44.8/5.7 | 48.5 / 5.3 | 57.8/3.8 |
| In Plane Rotation (IPR) | 52.1 / 4.1 | 50.7 / 5.1 | 46.4 / 5.3 | 45.7 / 5.9 | 43.7/5.9 | 47.1 / 5.5 | 50.7/4.9 |
| Occlusion (OCC) | 56.0 / 3.8 | 52.7 / 5.1 | 49.3 / 5.1 | 47.6 / 5.8 | 47.4/5.5 | 51.2 / 5.1 | 57.9/3.6 |
| Illumination Variation (IV) | 59.6 / 3.0 | 53.0 / 4.7 | 51.2 / 4.8 | 47.1 / 5.6 | 47.8/5.8 | 51.4 / 5.2 | 59.9/3.0 |

Table 8. Quantitative Results of Tracking under Occlusions

| Approaches | | ASLA | DFT | IVT | MIL | PCOM | LSS | Proposed |
|---|---|---|---|---|---|---|---|---|
| Average number of failures in 1000 frames under occlusion | Person A (red window) | 3.5 | 5.3 | 5.2 | 6.1 | 5.7 | 4.7 | 0.5 |
| | Person B (blue window) | 4.1 | 4.6 | 5.3 | 6.3 | 5.5 | 5.0 | 0.4 |
| | Person C (green window) | 3.8 | 5.4 | 4.8 | 5.0 | 5.3 | 5.6 | 0.6 |
| | Average | 3.8 | 5.1 | 5.1 | 5.8 | 5.5 | 5.1 | 0.5 |
| RMSE of Position (by pixels) | Person A (red window) | 12.734 | 11.245 | 9.363 | 8.342 | 8.632 | 13.297 | 2.642 |
| | Person B (blue window) | 5.246 | 7.147 | 5.754 | 5.329 | 5.387 | 10.298 | 2.167 |
| | Person C (green window) | 13.154 | 5.267 | 7.839 | 4.329 | 11.423 | 7.684 | 1.924 |
| | Average | 10.378 | 7.886 | 7.652 | 6.000 | 8.480 | 10.426 | 2.244 |

## V. Conclusion

This paper aims to simplify the detection, recognition and tracking of moving objects in real-time video via SP Theory of Intelligence and species inspired Particle Swarm Optimization (PSO). Wide variety of multiple alignments extracted from the video scenes of our object of interests are stored as Old information. Subsequently, on arrival of New information, the knowledge base is updated in order to derive optimum compressed pattern to follow in the successive frames. The unsupervised learning potentiality is explored in section III (C), where the system derives objects and classes of objects from its existing knowledge base in form of multiple alignment. As in case of human perception, the SP system is quite adaptive in the face of substitution, omission or commission. SP theory possesses a brain-like knowledge interface which is more robust and thorough over other intelligent systems. As briefly explained in this paper, with the help of multilevel abstraction, part-whole hierarchy, class hierarchy and polythetic concepts, we are getting breakthrough results in the field of vision and pattern recognition. For tracking purposes, species inspired PSO is more persistent in processing natural language based multiple alignments, which in turn produces more satisfactory results than other state-of-the-art artificial systems. This has great potentials in the field of problem solving integrating vision and pattern recognition with more robustness and variability, with exciting opportunities to explore in near future.